\documentclass[10pt,twocolumn,letterpaper]{article}

\usepackage{booktabs}      
\usepackage[table]{xcolor} 
\usepackage{tabularx}      

\usepackage[pagenumbers]{cvpr} 

\definecolor{cvprblue}{rgb}{0.21,0.49,0.74}
\usepackage[pagebackref,breaklinks,colorlinks,allcolors=cvprblue]{hyperref}

\def\confYear{2026}

\title{SkeleGuide: Explicit Skeleton Reasoning for Context-Aware Human-in-Place Image Synthesis}

\author{Chuqiao Wu\\
Alibaba Group\\
{\tt\small wuchuqiao.wcq@taobao.com}
\and
Jin Song\\
Alibaba Group\\
{\tt\small songjin.song@alibaba-inc.com}
\and
Yiyun Fei\\
Alibaba Group\\
{\tt\small yunhun.fyy@taobao.com}
}

\begin{document}
\maketitle
\begin{abstract}
Generating realistic and  structurally plausible human images into existing scenes remains a significant challenge for current generative models, which often produce artifacts like distorted limbs and unnatural poses. We attribute this systemic failure to an inability to perform explicit reasoning over human skeletal structure. To address this, we introduce SkeleGuide, a novel framework built upon explicit skeletal reasoning. Through joint training of its reasoning and rendering stages, SkeleGuide learns to produce an internal pose that acts as a strong structural prior, guiding the synthesis towards high structural integrity. For fine-grained user control, we introduce PoseInverter, a module that decodes this internal latent pose into an explicit and editable format.  Extensive experiments demonstrate that SkeleGuide significantly outperforms both specialized and general-purpose models in generating high-fidelity, contextually-aware human images. Our work provides compelling evidence that explicitly modeling skeletal structure is a fundamental step towards robust and plausible human image synthesis.
\end{abstract}    
\section{Introduction}
\label{sec:intro}

\begin{figure}[t]
  \centering
  \includegraphics[width=\columnwidth]{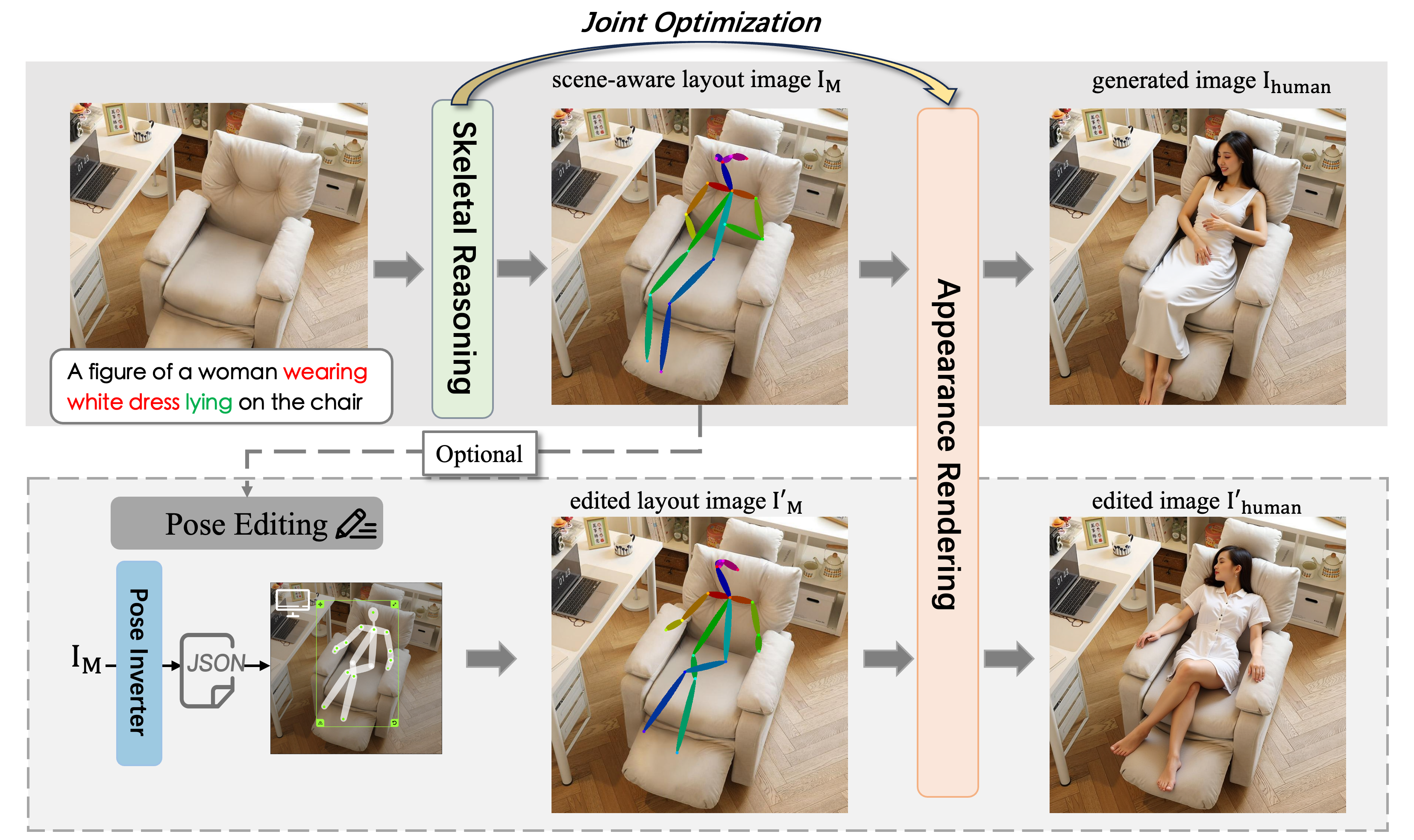}
  \caption{From just a scene image and a text prompt, SkeleGuide enables users to realistically place a person and precisely control the pose.}
  \label{fig:edit_flow}
\end{figure}

Recent advancements in diffusion models \cite{rombach2022high,flux2024,esser2024scaling} have revolutionized image generation, enabling the creation of highly realistic visual content from text prompts. A key challenge within this domain is Human-in-Place Synthesis: the task of realistically inserting a person into an existing scene (as illustrated in Fig.~\ref{fig:edit_flow}). 
Despite progress, state-of-the-art models often struggle with this, producing glaring artifacts like distorted limbs and unnatural poses that fall into the ``uncanny valley.'' These failures severely limit their utility in crucial applications like virtual try-on, advertising, and character design.

We contend that these issues are symptomatic of a fundamental limitation: the lack of an explicit mechanism for structural reasoning. The dominant generative paradigm, while effective for textures and global composition, fails to comprehend the human body as an articulated object governed by an underlying skeletal system. This skeleton acts as an ``anatomical grammar,'' a crucial piece of latent information dictating the valid configurations of body parts. Without a process to reason about this grammar, models are forced to learn these complex structural rules implicitly from pixel data alone—an extraordinarily difficult task that leads to the aforementioned inconsistencies. 
This limitation persists even in pose-guided human generation. While these methods use keypoints as explicit conditions, they typically treat the pose as a static template rather than the very structural grammar it is meant to represent, thus failing to enforce anatomical integrity.

A promising direction can be found in the paradigm shifts of Large Language Models (LLMs), particularly the concept of ``chain-of-thought'' (CoT) reasoning \cite{wei2022chain,
lampinen2022can,guo2025deepseek}. Prompting an LLM to generate intermediate reasoning steps before arriving at a final answer significantly improves its performance on complex tasks. We posit that a similar ``reason-then-render'' principle is key for human synthesis. To create a structurally coherent human image, a model must first ``think'' about its underlying structure. In this context, the ``thought'' is the explicit skeletal representation. By compelling the model to first conceptualize a skeleton, we ground the subsequent rendering process in a structurally sound foundation.

To this end, we introduce \textbf{SkeleGuide}, a framework built upon a ``reason-then-render'' paradigm (Fig.~\ref{fig:edit_flow}). As illustrated, it first employs a \textbf{Skeletal Reasoning} module to generate a scene-aware pose layout. This layout then directs a \textbf{Appearance Rendering} module to synthesize the final, high-fidelity image. While functionally distinct, the two modules are trained synergistically: the rendering stage provides supervisory feedback that regularizes the reasoning process, ensuring the generated layouts are not only plausible but also readily renderable.

This decoupled-yet-unified design unlocks two key advantages over contemporary methods like Person in Place \cite{yang2024person}, which employs a more fragmented, two-stage workflow. First, it allows for remarkable user convenience. Operating on only a scene image and a text prompt, SkeleGuide avoids the restrictive inputs like source person images or bounding boxes. Second, it makes the intermediate pose layout transparent and editable. As shown in Figure~\ref{fig:edit_flow}, our {Pose Inverter} module makes this internal structure accessible, allowing users to intuitively modify the pose. This transforms SkeleGuide from a simple generator into a versatile and controllable creative tool.

The contributions of our work are as follows:
\begin{itemize}
\item We introduce \textbf{SkeleGuide}, a novel framework for human-in-place image synthesis that unifies explicit skeletal reasoning with appearance rendering in a single pipeline. As the first framework to be jointly optimized in this manner, it establishes a new "reason-then-render" paradigm.

\item We propose the use of a \textbf{scene-aware layout} as a robust intermediate representation that effectively bridges the reasoning and rendering stages, which is the key to enabling our unified training strategy.

\item We achieve {new state-of-the-art performance} in human-in-place synthesis, with our method significantly outperforming prior specialized and general-purpose models in both quantitative and qualitative evaluations.
\end{itemize}

Our work demonstrates that integrating explicit skeletal reasoning into generative models is not merely an incremental improvement, but a foundational shift necessary for achieving true structural integrity and high-fidelity human synthesis.

\section{Related Work}
\label{sec:formatting}

\subsection{Controllable Diffusion Models}

The ascendancy of diffusion models stems from their formulation as denoising probabilistic frameworks \cite{ho2020denoising}, a principle that evolved into highly effective architectures like Latent Diffusion Models (LDM) \cite{rombach2022high}. A pivotal architectural shift has been the transition from traditional U-Net backbones to Diffusion Transformers (DiTs) \cite{peebles2023scalable}, as adopted by leading models like FLUX \cite{flux2024} and Stable Diffusion 3 \cite{esser2024scaling}. This has enabled remarkable gains in image quality, high-resolution synthesis, and the ability to process complex conditional inputs. 

To steer this power, a variety of control mechanisms have emerged, broadly categorized as: (i)~{spatially-aligned} methods that rely on explicit geometric cues (e.g., ControlNet~\cite{zhang2023adding}); (ii)~{semantically-aligned} approaches guided by reference images (e.g., IP-Adapter~\cite{ye2023ip}); and (iii)~{language-guided} editors, which span from instruction-following models~\cite{brooks2023instructpix2pix, DBLP:conf/nips/ZhangMCSS23, DBLP:conf/cvpr/0007YFQCYC0SEXX24, DBLP:conf/icml/NicholDRSMMSC22} to large multimodal models with editing capabilities like GPT-Image-1~\cite{hurst2024gpt} and Nano Banana~\cite{gemininanobanana}.
While these VLMs~\cite{ liu2025step1x-edit, labs2025flux1kontextflowmatching, wu2025qwen,DBLP:journals/tog/AvrahamiFL23}, empowered by massive-scale training, can alleviate some anatomical artifacts, they struggle with the precise task of \textbf{Human-in-Place Image Synthesis} by frequently altering the background and generating subjects with unnatural poses. Similarly, other recent efforts, such as Easycontrol and OminiControl \cite{zhang2025easycontrol, tan2024ominicontrol, huang2024context} have focused on adapting these controls to DiTs with parameter-efficient techniques like LoRA \cite{hu2022lora}. Ultimately, all these methods remain general-purpose tools and critically lack the specific structural priors necessary for generating anatomically plausible humans, a gap our work aims to fill.

\subsection{The Challenge of Human Image Synthesis}

Generating structurally plausible humans remains a formidable challenge. The field has explored several specialized strategies, each with inherent limitations. Bespoke generation models \cite{DBLP:conf/cvpr/NaKL25,jiang2022text2human} like CosmicMan \cite{li2024cosmicman}, Stylegan \cite{fu2022stylegan} achieve high quality but, as highly specialized systems, they lack the flexibility for versatile, in-context editing. Post-hoc refinement techniques such as HumanRefiner \cite{fang2024humanrefiner} act as corrective measures, fixing artifacts after generation rather than preventing them at the source. To directly address these structural issues, pose-guided image generation \cite{albahar2019guided,shen2023advancing,ju2023humansd,wang2024stable} has emerged as a prominent direction. The rationale for this approach is further reinforced by recent findings; Disentangled method \cite{xiao2025disentangled} highlight the importance of decoupling pose from appearance, while SP-Ctrl \cite{xuan2025rethink} show that sparse skeletal signals can achieve control comparable to dense guidance maps. Together, these insights establish skeletal structure as a powerful and efficient control modality. However, all such methods critically presuppose the availability of a clean target pose, a requirement often unmet in practical, text-driven editing scenarios.

\subsection{From Disjointed Pipelines to Integrated Reasoning}

Addressing this critical limitation—the need for a predefined pose—motivates a seemingly logical two-step approach: first predicting a plausible skeleton, then using it for guidance. The most relevant prior work, Person in Place \cite{yang2024person}, exemplifies this strategy. However, this approach suffers from several critical drawbacks. First, it imposes a heavy input burden by requiring not only a person box but also an object box, which limits its flexibility and ease of use. Second, its architecture is constrained to single-person generation per inference, lacking a mechanism for synthesizing coherent multi-person scenes in a single pass. Most fundamentally, its reliance on a disjointed pipeline—using a standalone skeleton generator to guide an entirely separate, off-the-shelf renderer—prevents end-to-end optimization and fails to capture the intricate interplay between structural planning and appearance.
\begin{figure*}[t]
\centering
\includegraphics[width=\textwidth]{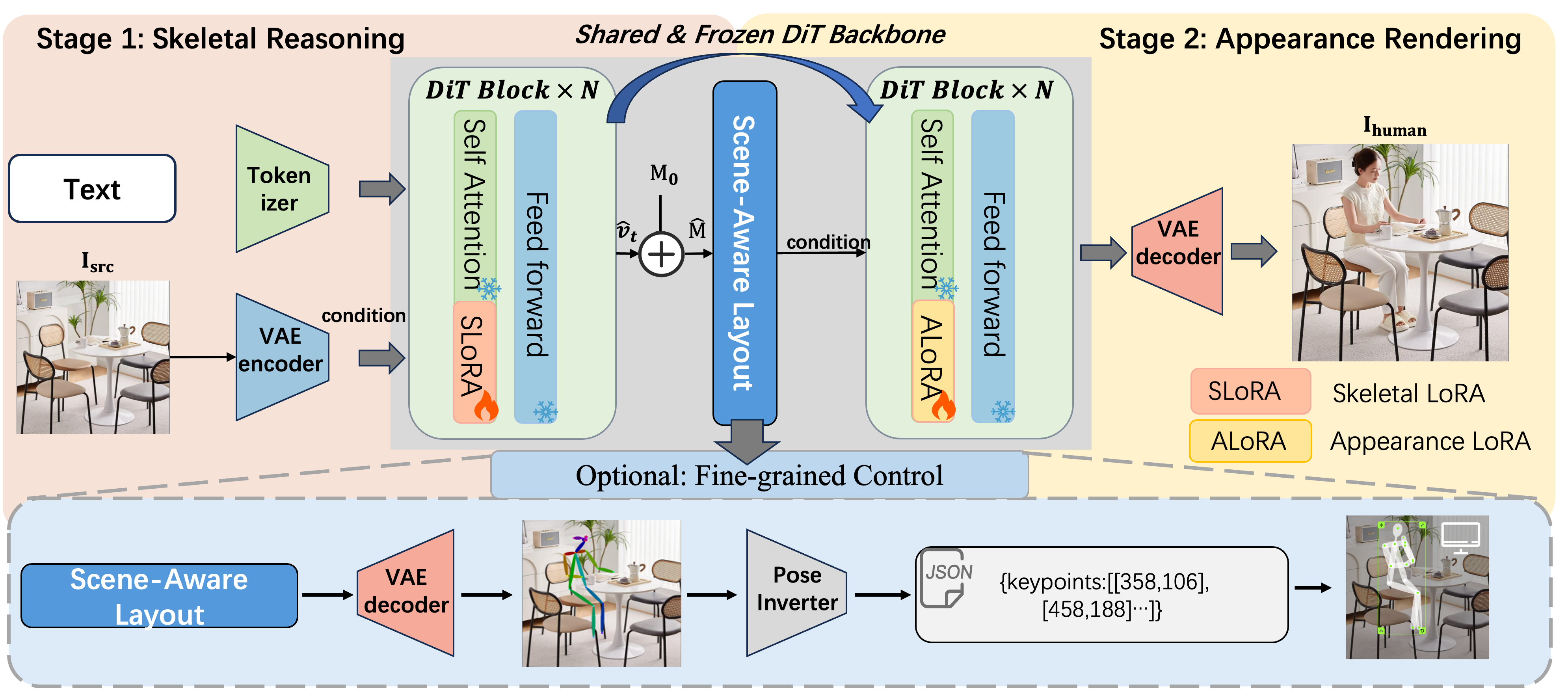} 
\caption{\textbf{Overview of our SkeleGuide framework.} Stage 1 (Skeletal Reasoning) generates a latent pose representation from text and a scene image. Stage 2 (Appearance Rendering) then synthesizes the final image conditioned on this latent pose. The optional control loop enables fine-grained editing of the intermediate pose.}

\label{fig1}
\end{figure*}

In stark contrast, SkeleGuide reframes the problem entirely. To our knowledge, we are the first to conceptualize intermediate skeleton generation not as a prerequisite task, but as an explicit reasoning step integral to the synthesis process itself. Our decoupled yet jointly trained  ``reason-then-render'' architecture learns to ``think'' structurally before it renders. This holistic approach allows our model to harmonize structure and appearance, systematically resolving the limitations of prior art and filling a critical gap in the literature.
\section{Method}

\subsection{Preliminarys}
\label{preliminarys}

Our approach is built upon the powerful combination of the Flow Matching (FM) training objective~\cite{lipman2022flow} and the Diffusion Transformer (DiT) architecture~\cite{peebles2023scalable}, a foundation popularized by recent state-of-the-art models~\cite{flux2024,esser2024scaling}.

Flow Matching (FM) learns a velocity field $\mathbf{v}$ that transforms a noise distribution $\mathbf{x}_0 \sim \mathcal{N}(0, \mathbf{I})$ into a data distribution $\mathbf{x}_1$. This is achieved by training a network $\mathbf{u}(\cdot)$ on a linear path $\mathbf{x}_t = (1-t)\mathbf{x}_0 + t\mathbf{x}_1$ for $t \in [0, 1]$ to predict the path's velocity vector, $\mathbf{v_t} = \mathbf{x}_1 - \mathbf{x}_0$. The stable training loss is:
\begin{equation}
    \mathcal{L}_{\text{FM}} = \mathbb{E}_{t, \mathbf{x}_0, \mathbf{x}_1, \mathbf{c}} \left[ \left\| \mathbf{u}(\mathbf{x}_t, t, \mathbf{c}; \theta) - (\mathbf{x}_1 - \mathbf{x}_0) \right\|_2^2 \right].
    \label{eq:fm_loss}
\end{equation}

The prediction network $\mathbf{u}$ in Eq.~\ref{eq:fm_loss} is often implemented using a Diffusion Transformer (DiT)~\cite{peebles2023scalable}. Unlike traditional U-Net architectures, a DiT operates on a sequence of latent patches, processing them through a series of Transformer blocks. For multi-modal conditioning, this architecture, often termed a Multi-Modal DiT (MM-DiT)~\cite{esser2024scaling,flux2024}, operates on a concatenated sequence of noisy image tokens $\mathbf{z}_t$ and condition tokens $\mathbf{c}$. This enables rich cross-modal interactions through a unified attention mechanism:
\begin{equation}
    \text{Attention}([\mathbf{z}_t, \mathbf{c}]) = \text{softmax}\left(\frac{\mathbf{QK}^\top}{\sqrt{d}}\right)\mathbf{V},
    \label{eq:attention}
\end{equation}
where $\mathbf{Q}, \mathbf{K}, \mathbf{V}$ are projected from the concatenated input.

\subsection{Overall Framework}

We address the task of \textbf{human-in-place image synthesis}: generating a realistic human image $\mathbf{I}_{\text{human}}$ within a given background scene $\mathbf{I}_{\text{src}}$ according to the prompt $\mathbf{T}$. A key challenge is maintaining anatomical correctness and avoiding artifacts like distorted limbs or unnatural poses. To tackle this, we introduce SkeleGuide, a two-stage framework that operationalizes a ``reason first, render second'' paradigm, as illustrated in Fig.~\ref{fig1}.

\textbf{Stage 1: Skeletal Reasoning.}
The Reasoning Module synthesizes the source image $\mathbf{I}_{\text{src}}$ and text prompt $\mathbf{T}$ into a predicted scene-aware layout, denoted as $\hat{\mathbf{M}}$. This latent representation is designed to serve as a holistic prior for the rendering stage by jointly encoding two critical elements: (1) an inferred skeletal structure that is contextually appropriate for the scene, and (2) the essential background context from $\mathbf{I}_{\text{src}}$. The resulting $\hat{\mathbf{M}}$ thus acts as a self-contained \textbf{structural and contextual plan}, providing all necessary guidance for the subsequent rendering module.

\textbf{Stage 2: Appearance Rendering.}
The Appearance Rendering stage then leverages this rich prior, $\hat{\mathbf{M}}$, and the original text prompt, $\mathbf{T}$, to synthesize the final image, $\mathbf{I}_{\text{human}}$. 
Its role is to render a human image that is seamlessly integrated with the scene by strictly adhering to the structural and contextual guidance encapsulated within $\hat{\mathbf{M}}$.


The synthesis of the scene-aware layout $\hat{\mathbf{M}}$ is the cornerstone of our two-stage design. Because the \textbf{generation process for $\hat{\mathbf{M}}$ is fully differentiable}, it is crucial for enabling end-to-end training. This allows gradients from the rendering loss to be backpropagated through the intermediate representation to update the reasoning stage. This process fosters co-adaptation, compelling the two stages to work in synergy. The decoupled architecture not only enhances robustness but also provides natural control points for fine-grained manipulation.

\subsection{Unified Architecture and Condition Injection}

A core design principle of SkeleGuide is the use of a \textbf{single, shared DiT backbone} for both its reasoning and rendering stages, promoting parameter efficiency. To specialize this backbone for its distinct roles, we employ a targeted LoRA injection strategy, applying Low-Rank Adaptation (LoRA)~\cite{hu2022lora} updates exclusively to the query, key, and value (QKV) projections of the \textbf{Condition Branch}~\cite{zhang2025easycontrol}. The Text and Noise branches, however, remain unmodified. This approach allows conditional information to be efficiently injected without disrupting the model's core pre-trained representations.

Specifically, for a given conditional token $\mathbf{z}_c$, the updated query feature $\mathbf{Q}'_c$ is computed by augmenting the original projection:
\begin{equation}
    \mathbf{Q}'_c = \mathbf{W}_Q \mathbf{z}_c + \mathbf{B}_Q \mathbf{A}_Q \mathbf{z}_c,
    \label{eq:lora_update}
\end{equation}
where $\mathbf{B}_Q \mathbf{A}_Q$ is the trainable low-rank update. The key and value features are updated analogously. This targeted approach allows conditional information to be efficiently injected without disrupting the model's core pre-trained representations.

We instantiate two unique sets of these QKV LoRA weights: a \textbf{Skeletal LoRA} for the reasoning stage and an \textbf{Appearance LoRA} for the rendering stage. Activating the appropriate set within the Condition Branch precisely steers the unified architecture to perform its designated role in our two-stage pipeline.

\subsection{Training and Inference Pipeline}

To train our ``reason-then-render'' pipeline effectively, we devise a progressive three-phase training strategy. This curriculum is designed to first establish a robust reasoning foundation, then foster synergy between the two modules, and finally polish the rendering quality for maximum fidelity.

\paragraph{Phase 1: Reasoning Module Pre-training.}

In the first phase, we train the Reasoning Module in isolation to predict a Scene-Aware Layout, denoted by its latent representation $\mathbf{M}$. To construct the ground-truth for this training, we first generate a target layout image, $\mathbf{I}_M$, by overlaying a ground-truth human skeleton—extracted via an off-the-shelf estimator~\cite{xu2022vitpose}—onto the source image. This image is then encoded into the target latent $\mathbf{M} = \mathcal{E}(\mathbf{I}_M)$ using a pre-trained VAE encoder $\mathcal{E}$.

The module is then trained with a \textbf{reasoning loss}, $\mathcal{L}_{\text{reason}}$, which follows the Flow Matching objective. This loss trains the network $\mathbf{u}_{\theta_1}$ to predict the velocity vector along a path from a noise latent $\mathbf{M}_0$ to the target latent $\mathbf{M}$:
\begin{equation}
    \mathcal{L}_{\text{reason}} = \mathbb{E} \left[ \left\| \mathbf{u}_{\theta_1}(\mathbf{M}_{t}, t, \mathbf{c}_{\text{src}}, \mathbf{c}_T) - (\mathbf{M} - \mathbf{M}_0) \right\|_2^2 \right],
    \label{eq:reason_loss}
\end{equation}
where $\mathbf{M}_t$ is the linear interpolation between $\mathbf{M}_0$ and $\mathbf{M}$. During this phase, we exclusively optimize the Skeletal LoRA weights.

\paragraph{Phase 2: Joint End-to-End Training.}

The second phase is crucial for fostering co-adaptation between the two modules. We achieve this by training them jointly with a combined objective, which introduces a \textbf{rendering loss}, $\mathcal{L}_{\text{render}}$. This loss guides the final image synthesis, conditioned on the predicted layout $\mathbf{c}_M$ from the first stage. To formulate this loss, we first encode the ground-truth human image, $\mathbf{I}_{\text{human}}$, into its target latent representation $\mathbf{z}_{h} = \mathcal{E}(\mathbf{I}_{\text{human}})$.

\begin{equation}
    \mathcal{L}_{\text{render}} = \mathbb{E} \left[ \left\| \mathbf{u}_{\theta_2}(\mathbf{z}_{h, t}, t, \mathbf{c}_T,  \mathbf{c}_M) - (\mathbf{z}_h - \mathbf{z}_{h, 0}) \right\|_2^2 \right],
    \label{eq:render_loss}
\end{equation}
where $\mathbf{z}_{h, t}$ is the interpolation between a noise latent $\mathbf{z}_{h, 0}$ and the ground-truth latent $\mathbf{z}_h$. The total loss is $\mathcal{L}_{\text{total}} = \lambda_{\text{reason}} \mathcal{L}_{\text{reason}} + \lambda_{\text{render}} \mathcal{L}_{\text{render}}$.

A critical aspect of this phase is how the layout condition $\mathbf{c}_M$ is derived. Instead of using the ground truth, we use the layout predicted online by the Reasoning Module. Specifically, given a pre-defined noise latent $\mathbf{M}_0$, the module predicts the velocity $\hat{\mathbf{v}} = \mathbf{u}_{\theta_1}(\cdot)$ required for denoising. As shown in Fig.~\ref{fig1}, the predicted clean scene-Aware layout, $\hat{\mathbf{M}}$, is then deterministically derived as:
\begin{equation}
    \hat{\mathbf{M}} = \mathbf{M}_0 + \hat{\mathbf{v}}_t.
    \label{eq:pose_derivation}
\end{equation}
This dynamically generated holistic prior $\hat{\mathbf{M}}$ is immediately used as the condition for the rendering loss (\emph{i.e.}, $\mathbf{c}_M = \hat{\mathbf{M}}$). This process mutually refines both modules: it trains the renderer to be robust to the characteristics of predicted layput, while compelling the reasoner to generate outputs that are more amenable to high-fidelity synthesis.

\paragraph{Phase 3: Rendering Module Fine-tuning.}
In the final phase, we refine the Rendering Module for maximum visual fidelity. The Reasoning Module's weights are frozen, and we optimize only the Appearance LoRA of the Rendering Module using just the rendering loss ($\mathcal{L}_{\text{total}} = \mathcal{L}_{\text{render}}$). Critically, and in contrast to Phase 2, the pose condition $\mathbf{c}_M$ is now sourced from the \textbf{ground-truth scene-aware layout}, \emph{i.e.}, $\mathbf{c}_M = \mathbf{M}$. By providing a ``perfect'' structural prior, this step allows the renderer to focus exclusively on translating accurate pose information into high-fidelity images.

\paragraph{Inference Pipeline.}
At inference time, the two stages operate as a clean, feed-forward cascade. First, given $\mathbf{I}_{\text{src}}$ and $\mathbf{T}$, the reasoning stage generates a scene-aware layout $\hat{\mathbf{M}}$. Crucially, the rendering stage then operates {exclusively} on this generated $\hat{\mathbf{M}}$ to produce the final image $\mathbf{I}_{\text{human}}$. This is possible because $\hat{\mathbf{M}}$ is designed to be a holistic representation, implicitly encoding not only the skeletal structure but also the essential appearance and background context from the original conditions. This design drastically simplifies the inference data flow and demonstrates the effectiveness of our reasoning stage in creating a self-contained, actionable plan for the renderer.

\subsection{Fine-Grained Pose Editing via Latent Inversion}
\label{sec:editing}

A key advantage of our explicit reasoning stage is the ability to enable fine-grained user control. To this end, we introduce a lightweight \textbf{{PoseInverter}} module, designed to make the model's {scene-aware layout}, $\hat{\mathbf{M}}$, transparent and editable.

The {PoseInverter} is a fine-tuned Vision-Language Model (VLM), QWEN-VL-2.5-7B ~\cite{Qwen2.5-VL}, trained to bridge the gap between visual images and structured JSON keypoints. Its training set consists of pairs of ground-truth visual images, denoted $\mathbf{I}_{M}$, and their corresponding JSON representations.

This enables a seamless interactive workflow. The latent representation $\hat{\mathbf{M}}$ generated by our model is first passed through a pre-trained VAE decoder to obtain its visual counterpart, $\hat{\mathbf{I}}_{M}$. This visual image is then fed into the {PoseInverter} to produce an editable JSON file. A user can manually adjust the keypoints in this file, and the modified structure $\hat{\mathbf{I}}^{'}_{M}$ is subsequently re-encoded back into a latent representation to guide the Appearance Rendering stage. This mechanism makes the model's internal structural plan directly accessible and manipulable, allowing for precise and intuitive control over the final pose.

\begin{table*}[t]
  \centering
  
  \renewcommand{\arraystretch}{0.9} 
  \setlength{\tabcolsep}{4pt} 
  
  \caption{\textbf{Quantitative evaluation on V-COCO.} We assess performance across three dimensions: full image quality, background consistency, and human generation quality. Our method, SkeleGuide, achieves state-of-the-art results in multiple key metrics. Best results are in \textbf{bold}.}
  \label{tab:quantitative_results_ultimate_final}
  
  \begin{tabular}{
    @{}
    p{3.9cm} 
    >{\centering\arraybackslash}p{1.1cm} 
    >{\centering\arraybackslash}p{1.1cm} 
    >{\centering\arraybackslash}p{1.1cm} 
    >{\centering\arraybackslash}p{1.1cm} 
    >{\centering\arraybackslash}p{1.1cm} 
    >{\centering\arraybackslash}p{1.1cm} 
    >{\centering\arraybackslash}p{1.8cm} 
    >{\centering\arraybackslash}p{1.5cm} 
    @{}
  }
    \toprule
    & \multicolumn{3}{c}{\textbf{Full Image Quality}} & \multicolumn{3}{c}{\textbf{Background Consistency}} & \multicolumn{2}{c}{\textbf{Human Generation Quality}} \\
    \cmidrule(lr){2-4} \cmidrule(lr){5-7} \cmidrule(lr){8-9}
    
    Method & FID$\,\downarrow$ & KID$\,\downarrow$ & CS$\,\uparrow$ & PSNR$\,\uparrow$ & SSIM$\,\uparrow$ & LPIPS$\,\downarrow$ & \mbox{ELO Rate}$\uparrow$ & HPSv2$\,\uparrow$ \\
    \midrule
    
    \rowcolor{gray!15}
    \multicolumn{9}{l}{\textit{Closed-source Commercial Models}} \\
    Gemini 2.0 Flash \cite{gemini2025flash} & 20.65 & 0.0027 & 25.32 & 20.95 & 0.685 & 0.193 &  983 & 0.193 \\

    GPT-Image-1 \cite{hurst2024gpt} & 23.18 & 0.0022 & \textbf{26.12} & 14.13 & 0.543  & 0.485 & 1089 & {0.221} \\
        Nano banana \cite{gemininanobanana} & 19.01 & 0.0016 & 25.20 & 19.11 & 0.607 & 0.260 & 1109 & 0.199 \\
    \midrule
    
    \rowcolor{gray!15}
    \multicolumn{9}{l}{\textit{Open-source General Models}} \\
    Step 1x-edit \cite{liu2025step1x-edit} & 18.49 & 0.0018 & {25.54} & 23.96 & 0.829 & 0.108 &  948 & 0.203 \\
    Flux-Kontext-dev \cite{labs2025flux1kontextflowmatching} &  26.85 & 0.0027 & 25.21 & 17.78 & 0.63 & 0.293 & 1026 & 0.219 \\
    Qwen-Image-Edit \cite{wu2025qwen} & 22.96 & 0.0018 & 25.51 & 19.37 & 0.669 &0.228 & 1036 & 0.204 \\
    \midrule

    \rowcolor{gray!15}
    \multicolumn{9}{l}{\textit{Open-source Dedicated Model}} \\
    Person in Place\cite{yang2024person} & 22.74 & 0.0021 & 25.27 & 20.65 & 0.694 & 0.178 &  689 &  0.173 \\
    \midrule
    
    \textbf{SkeleGuide (Ours)} & \textbf{15.48} & \textbf{0.0014} & 25.34 & \textbf{25.61} & \textbf{0.885} & \textbf{0.075} & \textbf{1120} & \textbf{0.227} \\
    
    \bottomrule
  \end{tabular}
\end{table*}
\section{Experiments}
This section details our experimental setup, including implementation, datasets, and evaluation metrics, followed by a comprehensive analysis of our results through qualitative, quantitative, and ablation studies.

\subsection{Implementation Details}\label{sec:implementation_details}
\paragraph{Training Settings.}SkeleGuide builds upon the pre-trained FLUX.1-dev~\cite{flux2024} DiT backbone. Both stages of our model are trained on a cluster of 8 NVIDIA H20 GPUs (96GB VRAM each) for over 10,000 steps per phase. We use a constant learning rate of $1 \times 10^{-4}$ with an effective batch size of 8 (1 per GPU). For the joint training in Phase 2, we set the loss weights to $\lambda_{\text{reason}} = 0.5$ and $\lambda_{\text{render}} = 1$. For inference, we employ a flow-matching sampler with 25 steps to generate the final images.

\paragraph{Dataset Construction.}

Our training set is a composite of the public V-COCO dataset~\cite{gupta2015visual} (7,698 images) and a self-curated home-decor dataset (6,581 images) designed to enhance data diversity. The creation process was twofold. First, we built the home-decor set by generating and manually refining high-fidelity (person, background) pairs. Second, we used these pairs to fine-tune a human removal model, which then automatically generated background-only counterparts for the entire V-COCO set. This data augmentation strategy is crucial for addressing the limitations of public datasets and achieving competitive performance, a standard practice in recent literature~\cite{zhang2025easycontrol, yang2024person,tan2024ominicontrol}. For full reproducibility, our pipeline is detailed in Supp. B.

\paragraph{Evaluation Metrics.}

We conduct a comprehensive evaluation across three critical dimensions, as detailed in Table~\ref{tab:quantitative_results_ultimate_final}.
 \textbf{Full Image Quality} is measured by standard perceptual metrics: FID~\cite{heusel2017gans} and KID~\cite{binkowski2018demystifying} for realism, and CLIP Score (CS)~\cite{hessel2021clipscore} for text alignment. 
 \textbf{Background Consistency} is assessed by computing PSNR, SSIM, and perceptual LPIPS~\cite{DBLP:conf/cvpr/ZhangIESW18} exclusively on the non-human regions, isolated via segmentation masks \cite{DBLP:journals/corr/abs-2401-03407}.
 \textbf{Human Generation Quality} is evaluated using two complementary approaches. First, we quantify anatomical correctness on the segmented human region using the Human Pose Score (HPSv2)~\cite{wu2023human}. Second, recognizing that automatic metrics often fail to capture subtle structural flaws, we conducted a comprehensive human preference study. 
  In this study, each of our five expert reviewers independently performed 8,400 blind pairwise comparisons — systematically evaluating all 28 unique method pairings (C(8,2)) for each of the 300 sampled prompts. From these judgments, where they selected the superior result for realism and structural integrity, we calculated win rates and derived a global Elo rating~\cite{elo1978rating} to capture overall perceptual quality.
 Full metric details are provided in the supplementary materials.

\subsection{Quantitative Results}

We evaluate all methods on the official V-COCO test set (3,318 images)~\cite{gupta2015visual}. To ensure a fair comparison, we standardized the evaluation protocol: all inputs were prepared by processing images with our human removal model, and a consistent prompt template was used for general-purpose editors. Since pre-processed data for Person in Place~\cite{yang2024person} was unavailable, we ran inference with their public model on our standardized set to obtain its results.

The quantitative results, presented in Table~\ref{tab:quantitative_results_ultimate_final}, demonstrate SkeleGuide's superior performance across all three key dimensions.
\textbf{(1) Full Image Quality:} Our method sets a new state-of-the-art in image realism, achieving the best FID (15.48) and KID (0.0014) scores by a significant margin. This demonstrates a clear advantage in generation quality, especially since text-to-image consistency (CS), a less challenging aspect, is comparable across all methods.
 \textbf{(2) Background Consistency:} Our method demonstrates exceptional background preservation, leading on all related metrics (PSNR, SSIM, LPIPS). This exposes a key weakness in certain general models like GPT-Image-1 \cite{hurst2024gpt}, which struggles with resolution mismatches and uncontrollable scene alterations, resulting in the poorest background fidelity (e.g., LPIPS 0.485).
\textbf{(3) Human Generation Quality:} 
SkeleGuide achieves the highest human preference Elo rating (1120), confirming its ability to generate perceptually plausible and aesthetically pleasing subjects. 
While large-scale commercial models also perform well due to their extensive training, our method's advantage is most evident when compared to {Person in Place} \cite{yang2024person}. 
Despite being a dedicated model, {Person in Place} scores lowest (689) because its decoupled pipeline frequently produces severe artifacts like distorted faces and twisted limbs. 
This highlights the critical role of our joint-training approach in ensuring structural integrity. 
SkeleGuide also obtains a competitive HPS score (0.227), further validating its anatomical correctness.


{In summary,} our results show that a specialized architecture can outperform even top-tier generalist models like \textbf{Nano banana \cite{gemininanobanana}} and \textbf{GPT-Image-1 \cite{hurst2024gpt}} on the targeted task of human-in-place image synthesis. SkeleGuide's superior balance of quality and efficiency underscores the continued value of purpose-built design in an era of large-scale models.

\begin{figure*}[h]
  \centering
  \includegraphics[width=\textwidth]{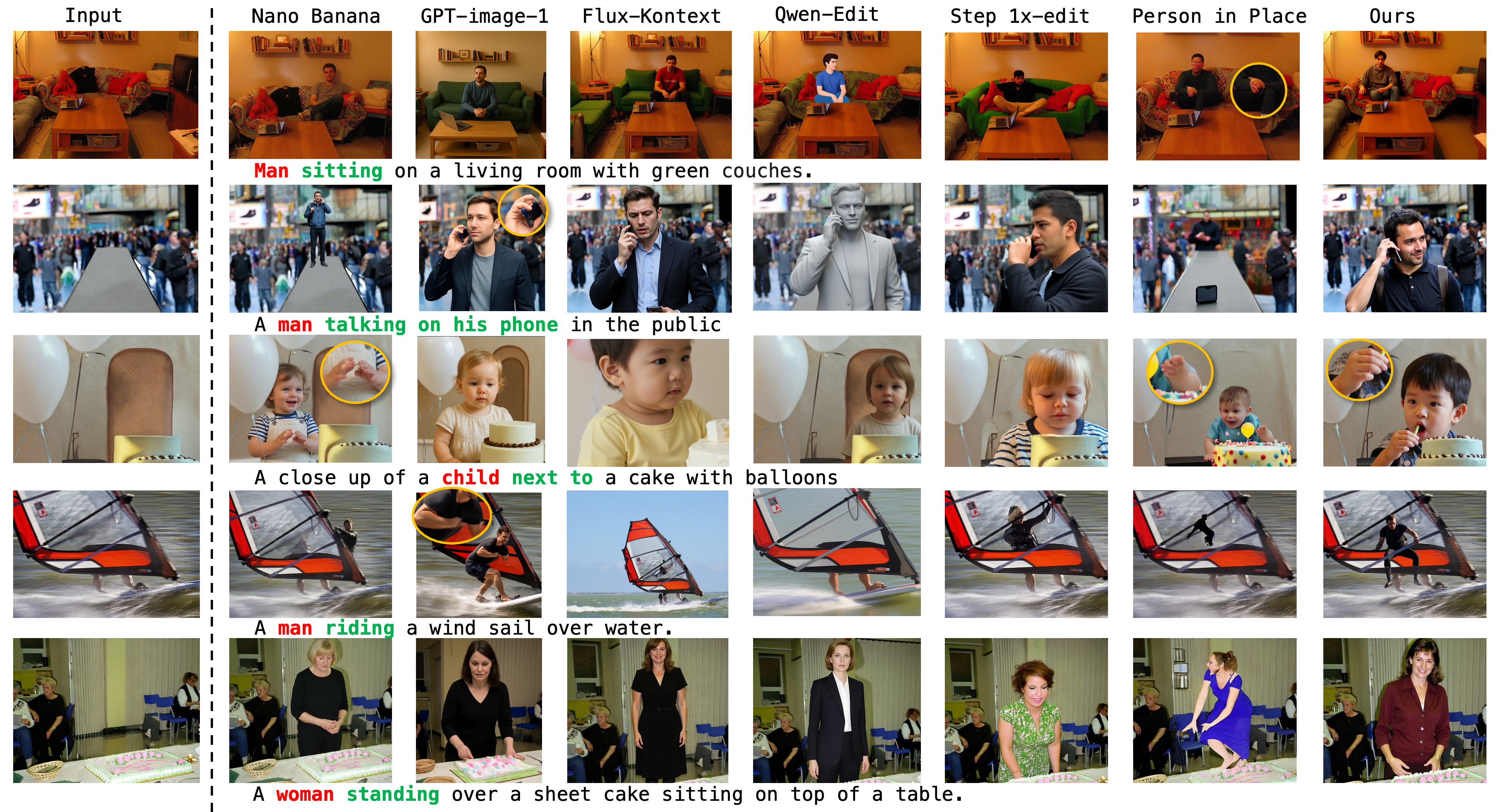}
  \caption{\textbf{General qualitative comparison with state-of-the-art methods.} 
    Across a variety of scenes, SkeleGuide demonstrates superior performance over specialized and general-purpose models.}
  \label{fig:qualitative_main}
\end{figure*}

\subsection{Qualitative Results}

Our qualitative analysis demonstrates SkeleGuide's superiority by first examining the quality of our reasoning stage, followed by a broader comparison with state-of-the-art models.

\begin{figure}[h]
  \centering
\includegraphics[width=\columnwidth]{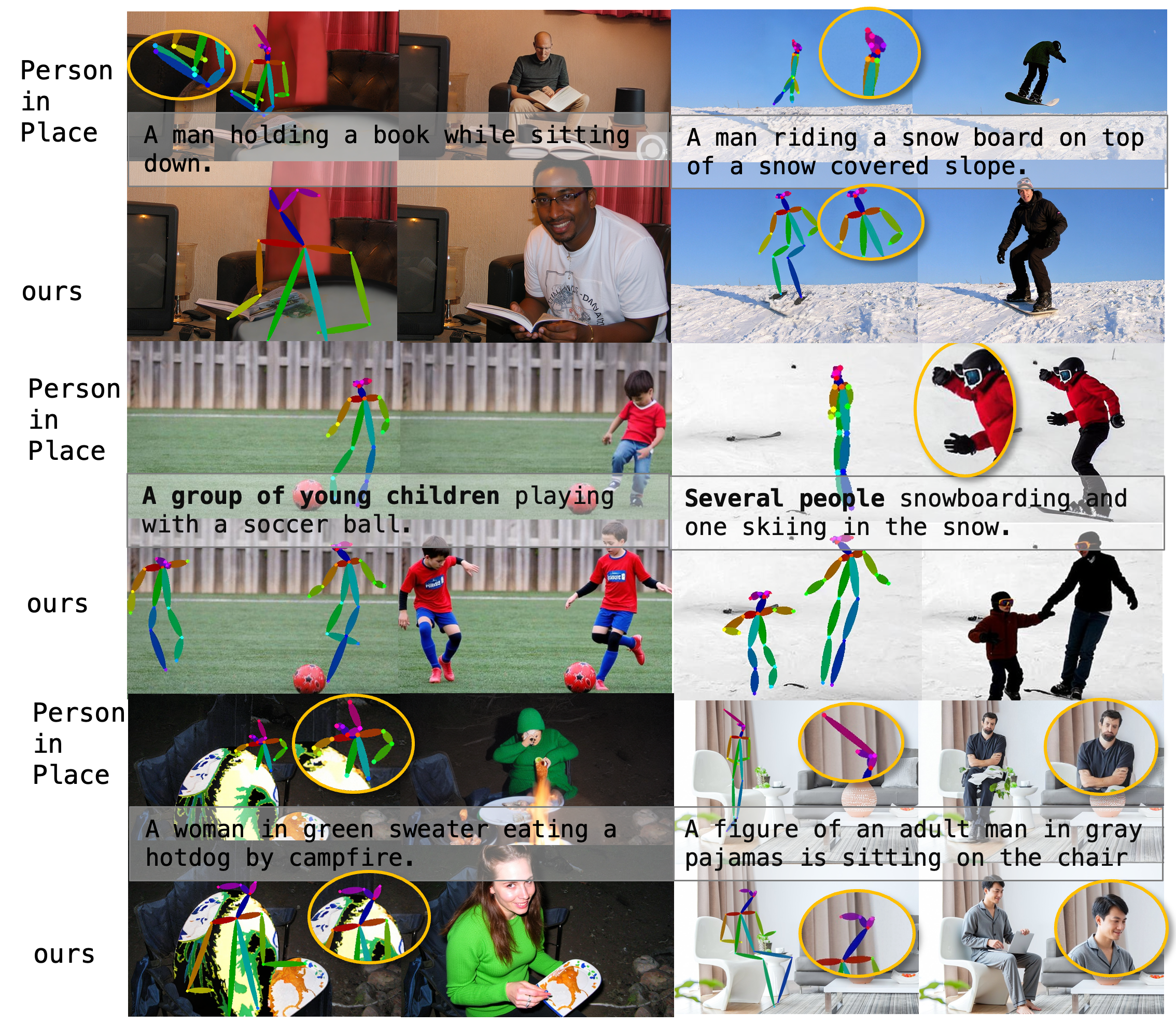}
  \caption{\textbf{Qualitative comparison with Person-in-Place.} SkeleGuide yields more coherent and plausible skeletons and final images with fewer artifacts.}
  \label{fig:qualitative_pose_compare}
\end{figure}

\paragraph{Quality of the Intermediate Pose Guidance.}
Figure~\ref{fig:qualitative_pose_compare} reveals that the inferiority of Person in Place~\cite{yang2024person} stems from its brittle architecture.Unlike our method which operates from minimal inputs, it requires restrictive conditions, including a source person and object bounding boxes, and employs a disjointed pipeline that decouples skeleton prediction from image synthesis. This decoupling frequently yields flawed intermediate poses with severely misplaced keypoints and tangled limbs (row 1, 3), resulting in physically impossible stances. Such foundational errors inevitably propagate, leading to distorted final images. Furthermore, this limitation is especially severe in multi-person scenarios. Its sequential, one-person-at-a-time inference process fundamentally fails to interpret group prompts, generating only a single subject or an incomplete scene (row 2).

In contrast, SkeleGuide's end-to-end architecture co-optimizes skeletal reasoning and image synthesis. This holistic approach ensures our generated poses are not only structurally sound and contextually aware but are also inherently synthesizable by our diffusion model. As a result, SkeleGuide robustly handles the complex single- and multi-person scenes where the restrictive and disjointed approach of competitors fails.

\paragraph{General Comparison with SOTA Editors.}

As illustrated in Figure \ref{fig:qualitative_main}, SkeleGuide excels in generating subjects with high structural fidelity while preserving background integrity, a capability where competing methods falter. The specialized model, Person in Place \cite{yang2024person}, and the general-purpose Step 1x-edit \cite{liu2025step1x-edit} both suffer from severe anatomical distortions, producing implausible poses and grotesque artifacts. Other general models, such as GPT-image-1 \cite{hurst2024gpt} and Flux-Kontext \cite{labs2025flux1kontextflowmatching}, are ill-suited for this in-place task, frequently destroying scene context by altering the entire background (row 1). Additional failure modes include the stylistic mismatch from Qwen-Edit \cite{wu2025qwen} (row 1, 2) and the imprecise placement and poor detail generation of Nano Banana \cite{gemininanobanana} (rows 2, 3, 4). In stark contrast, SkeleGuide consistently delivers realistic and contextually coherent results, successfully navigating the challenges that others fail to overcome.

\subsection{Ablation Study}
To validate the key design choices in SkeleGuide, we conduct a series of ablation studies. We investigate the effectiveness of our explicit, decoupled reasoning paradigm and the impact of the joint training strategy.

\paragraph{Effectiveness of Decoupled Reasoning.}
To validate the effectiveness of our explicit ``reason-then-render'' paradigm, we conduct an ablation study comparing SkeleGuide against two single-stage baselines, with results presented in Table~\ref{tab:ablation_decoupling}.

We compare SkeleGuide against two single-stage baselines. The first, a \textbf{naive single-stage} model, is trained to directly generate the final image from scene and text prompts, lacking any structural guidance. The second, a more advanced \textbf{implicit-reasoning} model, attempts to learn structure by predicting a concatenation of the final image and its corresponding pose map, forcing it to disentangle appearance and pose via self-attention alone.

  
  

\begin{table}[htbp]
  \centering
  \caption{\textbf{Ablation study on the reasoning paradigm.} We compare our explicit, two-stage approach against two single-stage baselines.}
  \label{tab:ablation_decoupling}

  \small
  \setlength{\tabcolsep}{3pt} 

  \begin{tabular*}{\columnwidth}{@{\extracolsep{\fill}}lcccccc@{}}
    \toprule
    Method & FID $\downarrow$ & KID $\downarrow$ & CS $\uparrow$ & PSNR $\uparrow$ & SSIM $\uparrow$ & LPIPS $\downarrow$ \\
    \midrule
    Single-Stage       & 19.63 & 0.0025 & 24.19 & 21.56 & 0.698 & 0.122 \\
    Implicit-Reas. & 18.59 & 0.0019 & 25.18 & 23.54 & 0.768 & 0.098 \\
    \midrule
    \textbf{SkeleGuide} & \textbf{15.48} & \textbf{0.0014} & \textbf{25.34} & \textbf{25.61} & \textbf{0.885} & \textbf{0.075} \\
    \bottomrule
  \end{tabular*}
\end{table}

The results in Table~\ref{tab:ablation_decoupling} clearly demonstrate the superiority of our explicit paradigm. The {Single-Stage} baseline, lacking any structural guidance, exhibits the poorest performance (FID 19.63). While adding an implicit objective in {Single-Stage(Implicit)} offers a marginal improvement (FID 18.59), it still falls significantly short of our explicit approach.

SkeleGuide achieves a substantially lower FID of {15.48} and SSIM of 0.885, validating our core hypothesis: explicitly generating an intermediate skeletal representation is fundamentally superior to relying on implicit, attention-based reasoning. This ``reason-then-render'' approach establishes a robust structural foundation, leading to a significant improvement in anatomical integrity and overall image quality.

\paragraph{Impact of Joint Training on Pose Guidance.}
To isolate the impact of joint training, we evaluate the quality of the intermediate scene-aware layout images generated by our full model versus a \textbf{Stage-1 Only} variant.


\begin{table}[htbp]
\centering
\caption{\textbf{Impact of joint training on pose guidance maps}, evaluated by PCK, preference rate (Pref.), PSNR, SSIM, and LPIPS.}
\label{tab:joint_training_impact_multicolumn}
\small
\setlength{\tabcolsep}{3pt} 

\begin{tabular*}{\columnwidth}{@{\extracolsep{\fill}}lccccc@{}} 
\toprule
Strategy & PCK@0.5 $\uparrow$ & Pref. $\uparrow$ & PSNR $\uparrow$ & SSIM $\uparrow$ & LPIPS $\downarrow$ \\
\midrule
Stage-1 Only & 31.05 & 38.33 & 27.15 & 0.814 & 0.068 \\ 
\textbf{Joint Training} & \textbf{54.05} & \textbf{61.53} & \textbf{29.32} & \textbf{0.920} & \textbf{0.057} \\ 
\bottomrule
\end{tabular*}
\end{table}

As shown in Table~\ref{tab:joint_training_impact_multicolumn}, joint training yields substantial improvements across all evaluation metrics. 
It not only boosts local keypoint accuracy PCK@0.5 score by over 23 points, but also excels in background consistency, leading on all PSNR, SSIM, and LPIPS. 
Crucially, it also dramatically improves the global structural coherence---an aspect that pixel-based metrics like PCK can fail to capture.
In a blind user study, our jointly trained model was preferred {61.53\%} of the time for this superior structural integrity.

\begin{figure}[h]
  \centering
  \includegraphics[width=\columnwidth]{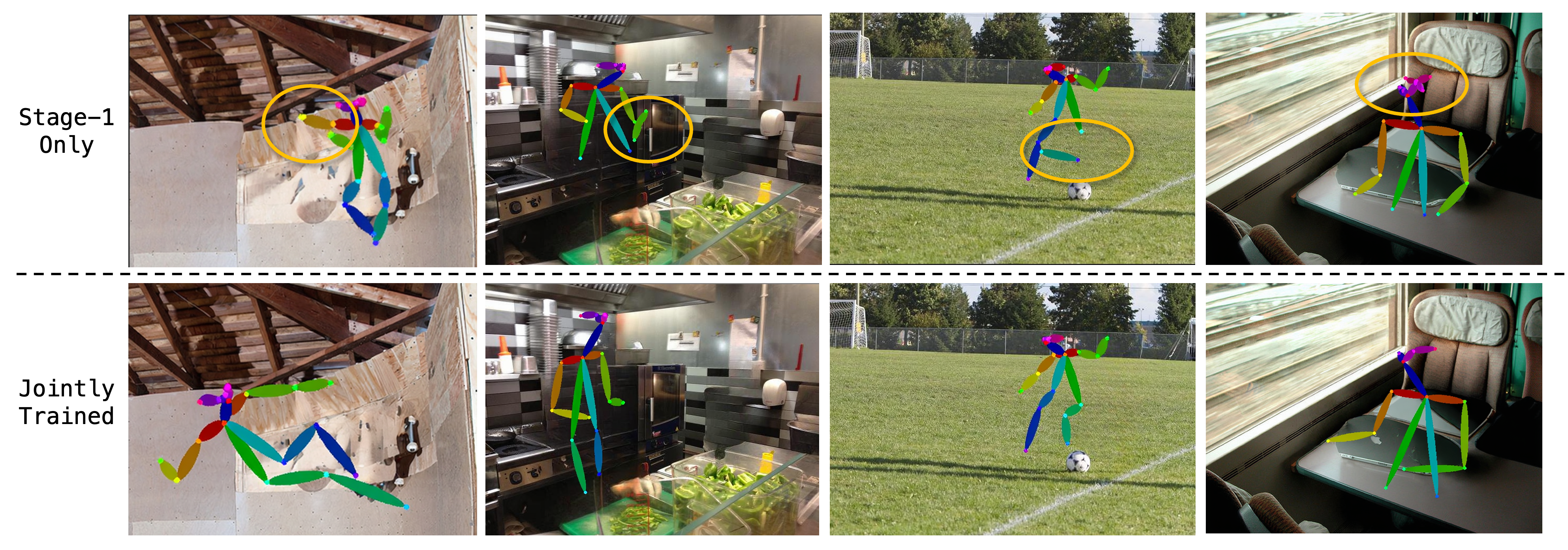}
  \caption{\textbf{Joint training enforces structural coherence.} Feedback from the rendering stage corrects the severe structural artifacts (e.g. incoherent limbs) produced when training Stage 1 alone.}
  \label{fig:ablation_joint_training}
\end{figure}

This quantitative improvement is visually confirmed in (Figure~\ref{fig:ablation_joint_training}). The {Stage-1 Only} variant frequently produces severe artifacts like floating limbs, whereas our full model generates consistently coherent and physically plausible poses.


\section{Conclusion}
In this paper, we presented SkeleGuide, a framework that significantly improves the structural integrity of synthesized human images. Our core contribution is a ``reason-then-render''  paradigm that introduces a {functional decoupling} within a single, {synergistically trained framework}. 
This design allows a reasoning stage to generate a robust layout, which is then refined by supervisory signals from the rendering stage, yielding state-of-the-art realism and consistency. 

The explicit layout representation also naturally affords fine-grained controllability, a capability we demonstrate through our PoseInverter module.
More broadly, our work advocates for structured, reasoning-based architectures as a compelling alternative to purely monolithic, end-to-end models, charting a promising path for future research in generative AI.

\clearpage
{
    \small
    \bibliographystyle{ieeenat_fullname}
    \bibliography{main}

@article{ho2020denoising,
  title={Denoising diffusion probabilistic models},
  author={Ho, Jonathan and Jain, Ajay and Abbeel, Pieter},
  journal={Advances in neural information processing systems},
  volume={33},
  pages={6840--6851},
  year={2020}
}

@inproceedings{rombach2022high,
  title={High-resolution image synthesis with latent diffusion models},
  author={Rombach, Robin and Blattmann, Andreas and Lorenz, Dominik and Esser, Patrick and Ommer, Bj{\"o}rn},
  booktitle={Proceedings of the IEEE/CVF conference on computer vision and pattern recognition},
  pages={10684--10695},
  year={2022}
}

@inproceedings{esser2024scaling,
  title={Scaling rectified flow transformers for high-resolution image synthesis},
  author={Esser, Patrick and Kulal, Sumith and Blattmann, Andreas and Entezari, Rahim and M{\"u}ller, Jonas and Saini, Harry and Levi, Yam and Lorenz, Dominik and Sauer, Axel and Boesel, Frederic and others},
  booktitle={Forty-first international conference on machine learning},
  year={2024}
}

@inproceedings{brooks2023instructpix2pix,
  title={Instructpix2pix: Learning to follow image editing instructions},
  author={Brooks, Tim and Holynski, Aleksander and Efros, Alexei A},
  booktitle={Proceedings of the IEEE/CVF conference on computer vision and pattern recognition},
  pages={18392--18402},
  year={2023}
}

@article{tan2024ominicontrol,
  title={Ominicontrol: Minimal and universal control for diffusion transformer},
  author={Tan, Zhenxiong and Liu, Songhua and Yang, Xingyi and Xue, Qiaochu and Wang, Xinchao},
  journal={arXiv preprint arXiv:2411.15098},
  year={2024}
}

@article{zhang2025easycontrol,
  title={Easycontrol: Adding efficient and flexible control for diffusion transformer},
  author={Zhang, Yuxuan and Yuan, Yirui and Song, Yiren and Wang, Haofan and Liu, Jiaming},
  journal={arXiv preprint arXiv:2503.07027},
  year={2025}
}

@inproceedings{DBLP:conf/cvpr/NaKL25,
  author       = {Sanghyeon Na and
                  Yonggyu Kim and
                  Hyunjoon Lee},
  title        = {Boost Your Human Image Generation Model via Direct Preference Optimization},
  booktitle    = {{IEEE/CVF} Conference on Computer Vision and Pattern Recognition,
                  {CVPR} 2025, Nashville, TN, USA, June 11-15, 2025},
  pages        = {23551--23562},
  publisher    = {Computer Vision Foundation / {IEEE}},
  year         = {2025},
  url          = {https://openaccess.thecvf.com/content/CVPR2025/html/Na\_Boost\_Your\_Human\_Image\_Generation\_Model\_via\_Direct\_Preference\_Optimization\_CVPR\_2025\_paper.html},
  doi          = {10.1109/CVPR52734.2025.02193},
  bibsource    = {dblp computer science bibliography, https://dblp.org}
}

@misc{flux2024,
    author={Black Forest Labs},
    title={FLUX},
    year={2024},
}

@article{DBLP:journals/tog/AvrahamiFL23,
  author       = {Omri Avrahami and
                  Ohad Fried and
                  Dani Lischinski},
  title        = {Blended Latent Diffusion},
  journal      = {{ACM} Trans. Graph.},
  volume       = {42},
  number       = {4},
  pages        = {149:1--149:11},
  year         = {2023},
  url          = {https://doi.org/10.1145/3592450},
  doi          = {10.1145/3592450},
  bibsource    = {dblp computer science bibliography, https://dblp.org}
}

@inproceedings{DBLP:conf/icml/NicholDRSMMSC22,
  author       = {Alexander Quinn Nichol and
                  Prafulla Dhariwal and
                  Aditya Ramesh and
                  Pranav Shyam and
                  Pamela Mishkin and
                  Bob McGrew and
                  Ilya Sutskever and
                  Mark Chen},
  editor       = {Kamalika Chaudhuri and
                  Stefanie Jegelka and
                  Le Song and
                  Csaba Szepesv{\'{a}}ri and
                  Gang Niu and
                  Sivan Sabato},
  title        = {{GLIDE:} Towards Photorealistic Image Generation and Editing with
                  Text-Guided Diffusion Models},
  booktitle    = {International Conference on Machine Learning, {ICML} 2022, 17-23 July
                  2022, Baltimore, Maryland, {USA}},
  series       = {Proceedings of Machine Learning Research},
  volume       = {162},
  pages        = {16784--16804},
  publisher    = {{PMLR}},
  year         = {2022},
  url          = {https://proceedings.mlr.press/v162/nichol22a.html},
  bibsource    = {dblp computer science bibliography, https://dblp.org}
}

@inproceedings{DBLP:conf/cvpr/0007YFQCYC0SEXX24,
  author       = {Shu Zhang and
                  Xinyi Yang and
                  Yihao Feng and
                  Can Qin and
                  Chia{-}Chih Chen and
                  Ning Yu and
                  Zeyuan Chen and
                  Huan Wang and
                  Silvio Savarese and
                  Stefano Ermon and
                  Caiming Xiong and
                  Ran Xu},
  title        = {{HIVE:} Harnessing Human Feedback for Instructional Visual Editing},
  booktitle    = {{IEEE/CVF} Conference on Computer Vision and Pattern Recognition,
                  {CVPR} 2024, Seattle, WA, USA, June 16-22, 2024},
  pages        = {9026--9036},
  publisher    = {{IEEE}},
  year         = {2024},
  url          = {https://doi.org/10.1109/CVPR52733.2024.00862},
  doi          = {10.1109/CVPR52733.2024.00862},
  bibsource    = {dblp computer science bibliography, https://dblp.org}
}

@inproceedings{DBLP:conf/nips/ZhangMCSS23,
  author       = {Kai Zhang and
                  Lingbo Mo and
                  Wenhu Chen and
                  Huan Sun and
                  Yu Su},
  editor       = {Alice Oh and
                  Tristan Naumann and
                  Amir Globerson and
                  Kate Saenko and
                  Moritz Hardt and
                  Sergey Levine},
  title        = {MagicBrush: {A} Manually Annotated Dataset for Instruction-Guided
                  Image Editing},
  booktitle    = {Advances in Neural Information Processing Systems 36: Annual Conference
                  on Neural Information Processing Systems 2023, NeurIPS 2023, New Orleans,
                  LA, USA, December 10 - 16, 2023},
  year         = {2023},
  url          = {http://papers.nips.cc/paper\_files/paper/2023/hash/64008fa30cba9b4d1ab1bd3bd3d57d61-Abstract-Datasets\_and\_Benchmarks.html},
  bibsource    = {dblp computer science bibliography, https://dblp.org}
}

@article{DBLP:journals/corr/abs-2401-03407,
  author       = {Peng Zheng and
                  Dehong Gao and
                  Deng{-}Ping Fan and
                  Li Liu and
                  Jorma Laaksonen and
                  Wanli Ouyang and
                  Nicu Sebe},
  title        = {Bilateral Reference for High-Resolution Dichotomous Image Segmentation},
  journal      = {CoRR},
  volume       = {abs/2401.03407},
  year         = {2024},
  url          = {https://doi.org/10.48550/arXiv.2401.03407},
  doi          = {10.48550/ARXIV.2401.03407},
  eprinttype    = {arXiv},
  eprint       = {2401.03407},
  bibsource    = {dblp computer science bibliography, https://dblp.org}
}

@inproceedings{peebles2023scalable,
  title={Scalable diffusion models with transformers},
  author={Peebles, William and Xie, Saining},
  booktitle={Proceedings of the IEEE/CVF international conference on computer vision},
  pages={4195--4205},
  year={2023}
}

@inproceedings{zhang2023adding,
  title={Adding conditional control to text-to-image diffusion models},
  author={Zhang, Lvmin and Rao, Anyi and Agrawala, Maneesh},
  booktitle={Proceedings of the IEEE/CVF international conference on computer vision},
  pages={3836--3847},
  year={2023}
}

@article{ye2023ip,
  title={Ip-adapter: Text compatible image prompt adapter for text-to-image diffusion models},
  author={Ye, Hu and Zhang, Jun and Liu, Sibo and Han, Xiao and Yang, Wei},
  journal={arXiv preprint arXiv:2308.06721},
  year={2023}
}

@article{huang2024context,
  title={In-context lora for diffusion transformers},
  author={Huang, Lianghua and Wang, Wei and Wu, Zhi-Fan and Shi, Yupeng and Dou, Huanzhang and Liang, Chen and Feng, Yutong and Liu, Yu and Zhou, Jingren},
  journal={arXiv preprint arXiv:2410.23775},
  year={2024}
}

@article{hu2022lora,
  title={Lora: Low-rank adaptation of large language models.},
  author={Hu, Edward J and Shen, Yelong and Wallis, Phillip and Allen-Zhu, Zeyuan and Li, Yuanzhi and Wang, Shean and Wang, Lu and Chen, Weizhu and others},
  journal={ICLR},
  volume={1},
  number={2},
  pages={3},
  year={2022}
}

@inproceedings{li2024cosmicman,
  title={Cosmicman: A text-to-image foundation model for humans},
  author={Li, Shikai and Fu, Jianglin and Liu, Kaiyuan and Wang, Wentao and Lin, Kwan-Yee and Wu, Wayne},
  booktitle={Proceedings of the IEEE/CVF Conference on Computer Vision and Pattern Recognition},
  pages={6955--6965},
  year={2024}
}

@inproceedings{fang2024humanrefiner,
  title={Humanrefiner: Benchmarking abnormal human generation and refining with coarse-to-fine pose-reversible guidance},
  author={Fang, Guian and Yan, Wenbiao and Guo, Yuanfan and Han, Jianhua and Jiang, Zutao and Xu, Hang and Liao, Shengcai and Liang, Xiaodan},
  booktitle={European Conference on Computer Vision},
  pages={201--217},
  year={2024},
  organization={Springer}
}

@inproceedings{ju2023humansd,
  title={Humansd: A native skeleton-guided diffusion model for human image generation},
  author={Ju, Xuan and Zeng, Ailing and Zhao, Chenchen and Wang, Jianan and Zhang, Lei and Xu, Qiang},
  booktitle={Proceedings of the IEEE/CVF International Conference on Computer Vision},
  pages={15988--15998},
  year={2023}
}

@article{wang2024stable,
  title={Stable-pose: Leveraging transformers for pose-guided text-to-image generation},
  author={Wang, Jiajun and Ghahremani Boozandani, Morteza and Li, Yitong and Ommer, Bj{\"o}rn and Wachinger, Christian},
  journal={Advances in Neural Information Processing Systems},
  volume={37},
  pages={65670--65698},
  year={2024}
}

@inproceedings{yang2024person,
  title={Person in Place: Generating Associative Skeleton-Guidance Maps for Human-Object Interaction Image Editing},
  author={Yang, ChangHee and Kang, ChanHee and Kong, Kyeongbo and Oh, Hanni and Kang, Suk-Ju},
  booktitle={Proceedings of the IEEE/CVF Conference on Computer Vision and Pattern Recognition},
  pages={8164--8175},
  year={2024}
}

@article{lipman2022flow,
  title={Flow matching for generative modeling},
  author={Lipman, Yaron and Chen, Ricky TQ and Ben-Hamu, Heli and Nickel, Maximilian and Le, Matt},
  journal={arXiv preprint arXiv:2210.02747},
  year={2022}
}

@article{xu2022vitpose,
  title={Vitpose: Simple vision transformer baselines for human pose estimation},
  author={Xu, Yufei and Zhang, Jing and Zhang, Qiming and Tao, Dacheng},
  journal={Advances in neural information processing systems},
  volume={35},
  pages={38571--38584},
  year={2022}
}

@article{heusel2017gans,
  title={Gans trained by a two time-scale update rule converge to a local nash equilibrium},
  author={Heusel, Martin and Ramsauer, Hubert and Unterthiner, Thomas and Nessler, Bernhard and Hochreiter, Sepp},
  journal={Advances in neural information processing systems},
  volume={30},
  year={2017}
}

@article{binkowski2018demystifying,
  title={Demystifying mmd gans},
  author={Bi{\'n}kowski, Miko{\l}aj and Sutherland, Danica J and Arbel, Michael and Gretton, Arthur},
  journal={arXiv preprint arXiv:1801.01401},
  year={2018}
}

@article{hessel2021clipscore,
  title={Clipscore: A reference-free evaluation metric for image captioning},
  author={Hessel, Jack and Holtzman, Ari and Forbes, Maxwell and Bras, Ronan Le and Choi, Yejin},
  journal={arXiv preprint arXiv:2104.08718},
  year={2021}
}

@article{gupta2015visual,
  title={Visual semantic role labeling},
  author={Gupta, Saurabh and Malik, Jitendra},
  journal={arXiv preprint arXiv:1505.04474},
  year={2015}
}

@inproceedings{elo1978rating,
  title={The rating of chessplayers, past and present},
  author={Arpad E. Elo},
  year={1978},
  url={https://api.semanticscholar.org/CorpusID:142610973}
}

@inproceedings{albahar2019guided,
  title={Guided image-to-image translation with bi-directional feature transformation},
  author={AlBahar, Badour and Huang, Jia-Bin},
  booktitle={Proceedings of the IEEE/CVF international conference on computer vision},
  pages={9016--9025},
  year={2019}
}

@article{jiang2022text2human,
  title={Text2human: Text-driven controllable human image generation},
  author={Jiang, Yuming and Yang, Shuai and Qiu, Haonan and Wu, Wayne and Loy, Chen Change and Liu, Ziwei},
  journal={ACM Transactions on Graphics (TOG)},
  volume={41},
  number={4},
  pages={1--11},
  year={2022},
  publisher={ACM New York, NY, USA}
}

@inproceedings{fu2022stylegan,
  title={Stylegan-human: A data-centric odyssey of human generation},
  author={Fu, Jianglin and Li, Shikai and Jiang, Yuming and Lin, Kwan-Yee and Qian, Chen and Loy, Chen Change and Wu, Wayne and Liu, Ziwei},
  booktitle={European Conference on Computer Vision},
  pages={1--19},
  year={2022},
  organization={Springer}
}

@article{shen2023advancing,
  title={Advancing pose-guided image synthesis with progressive conditional diffusion models},
  author={Shen, Fei and Ye, Hu and Zhang, Jun and Wang, Cong and Han, Xiao and Yang, Wei},
  journal={arXiv preprint arXiv:2310.06313},
  year={2023}
}

@article{Qwen2.5-VL,
  title={Qwen2.5-VL Technical Report},
  author={Bai, Shuai and Chen, Keqin and Liu, Xuejing and Wang, Jialin and Ge, Wenbin and Song, Sibo and Dang, Kai and Wang, Peng and Wang, Shijie and Tang, Jun and Zhong, Humen and Zhu, Yuanzhi and Yang, Mingkun and Li, Zhaohai and Wan, Jianqiang and Wang, Pengfei and Ding, Wei and Fu, Zheren and Xu, Yiheng and Ye, Jiabo and Zhang, Xi and Xie, Tianbao and Cheng, Zesen and Zhang, Hang and Yang, Zhibo and Xu, Haiyang and Lin, Junyang},
  journal={arXiv preprint arXiv:2502.13923},
  year={2025}
}

@article{liu2025step1x-edit,
      title={Step1X-Edit: A Practical Framework for General Image Editing}, 
      author={Shiyu Liu and Yucheng Han and Peng Xing and Fukun Yin and Rui Wang and Wei Cheng and Jiaqi Liao and Yingming Wang and Honghao Fu and Chunrui Han and Guopeng Li and Yuang Peng and Quan Sun and Jingwei Wu and Yan Cai and Zheng Ge and Ranchen Ming and Lei Xia and Xianfang Zeng and Yibo Zhu and Binxing Jiao and Xiangyu Zhang and Gang Yu and Daxin Jiang},
      journal={arXiv preprint arXiv:2504.17761},
      year={2025}
}

@misc{labs2025flux1kontextflowmatching,
      title={FLUX.1 Kontext: Flow Matching for In-Context Image Generation and Editing in Latent Space}, 
      author={Black Forest Labs and Stephen Batifol and Andreas Blattmann and Frederic Boesel and Saksham Consul and Cyril Diagne and Tim Dockhorn and Jack English and Zion English and Patrick Esser and Sumith Kulal and Kyle Lacey and Yam Levi and Cheng Li and Dominik Lorenz and Jonas Müller and Dustin Podell and Robin Rombach and Harry Saini and Axel Sauer and Luke Smith},
      year={2025},
      eprint={2506.15742},
      archivePrefix={arXiv},
      primaryClass={cs.GR},
      url={https://arxiv.org/abs/2506.15742},
}

@article{wu2025qwen,
  title={Qwen-image technical report},
  author={Wu, Chenfei and Li, Jiahao and Zhou, Jingren and Lin, Junyang and Gao, Kaiyuan and Yan, Kun and Yin, Sheng-ming and Bai, Shuai and Xu, Xiao and Chen, Yilei and others},
  journal={arXiv preprint arXiv:2508.02324},
  year={2025}
}

@inproceedings{DBLP:conf/cvpr/ZhangIESW18,
  author       = {Richard Zhang and
                  Phillip Isola and
                  Alexei A. Efros and
                  Eli Shechtman and
                  Oliver Wang},
  title        = {The Unreasonable Effectiveness of Deep Features as a Perceptual Metric},
  booktitle    = {2018 {IEEE} Conference on Computer Vision and Pattern Recognition,
                  {CVPR} 2018, Salt Lake City, UT, USA, June 18-22, 2018},
  pages        = {586--595},
  publisher    = {Computer Vision Foundation / {IEEE} Computer Society},
  year         = {2018},
  url          = {http://openaccess.thecvf.com/content\_cvpr\_2018/html/Zhang\_The\_Unreasonable\_Effectiveness\_CVPR\_2018\_paper.html},
  doi          = {10.1109/CVPR.2018.00068},
  bibsource    = {dblp computer science bibliography, https://dblp.org}
}

@misc{gemini2025flash,
  author       = {{Gemini Team Google}},
  title        = {{Gemini 2.0 Flash}},

  year         = {2025},
  month        = {Feb},
  note         = {Accessed: 2024-02-5}
}

@misc{gemininanobanana,
  author       = {{Gemini Team Google}},
  title        = {{Gemini 2.5 Flash}},
  
  year         = {2025},
  month        = {Aug},
  note         = {Accessed: 2025-08-26}
}

@article{hurst2024gpt,
  title={Gpt-4o system card},
  author={Hurst, Aaron and Lerer, Adam and Goucher, Adam P and Perelman, Adam and Ramesh, Aditya and Clark, Aidan and Ostrow, AJ and Welihinda, Akila and Hayes, Alan and Radford, Alec and others},
  journal={arXiv preprint arXiv:2410.21276},
  year={2024}
}

@article{wei2022chain,
  title={Chain-of-thought prompting elicits reasoning in large language models},
  author={Wei, Jason and Wang, Xuezhi and Schuurmans, Dale and Bosma, Maarten and Xia, Fei and Chi, Ed and Le, Quoc V and Zhou, Denny and others},
  journal={Advances in neural information processing systems},
  volume={35},
  pages={24824--24837},
  year={2022}
}

@article{lampinen2022can,
  title={Can language models learn from explanations in context?},
  author={Lampinen, Andrew K and Dasgupta, Ishita and Chan, Stephanie CY and Matthewson, Kory and Tessler, Michael Henry and Creswell, Antonia and McClelland, James L and Wang, Jane X and Hill, Felix},
  journal={arXiv preprint arXiv:2204.02329},
  year={2022}
}

@article{wu2023human,
  title={Human preference score v2: A solid benchmark for evaluating human preferences of text-to-image synthesis},
  author={Wu, Xiaoshi and Hao, Yiming and Sun, Keqiang and Chen, Yixiong and Zhu, Feng and Zhao, Rui and Li, Hongsheng},
  journal={arXiv preprint arXiv:2306.09341},
  year={2023}
}

@inproceedings{xiao2025disentangled,
  title={Disentangled Pose and Appearance Guidance for Multi-Pose Generation},
  author={Xiao, Tengfei and Wu, Yue and Li, Yuelong and Qin, Can and Gong, Maoguo and Miao, Qiguang and Ma, Wenping},
  booktitle={Proceedings of the Computer Vision and Pattern Recognition Conference},
  pages={5646--5655},
  year={2025}
}

@article{xuan2025rethink,
  title={Rethink Sparse Signals for Pose-guided Text-to-image Generation},
  author={Xuan, Wenjie and Zhang, Jing and Liu, Juhua and Du, Bo and Tao, Dacheng},
  journal={arXiv preprint arXiv:2506.20983},
  year={2025}
}

@article{guo2025deepseek,
  title={Deepseek-r1: Incentivizing reasoning capability in llms via reinforcement learning},
  author={Guo, Daya and Yang, Dejian and Zhang, Haowei and Song, Junxiao and Zhang, Ruoyu and Xu, Runxin and Zhu, Qihao and Ma, Shirong and Wang, Peiyi and Bi, Xiao and others},
  journal={arXiv preprint arXiv:2501.12948},
  year={2025}
}
}

\end{document}